%% file: main.tex
\documentclass[10pt,twocolumn,letterpaper]{article}

 \usepackage{cvpr}              

\input{preamble}

\definecolor{cvprblue}{rgb}{0.21,0.49,0.74}
\usepackage[pagebackref,breaklinks,colorlinks,citecolor=cvprblue]{hyperref}

\def\paperID{22} 
\def\confName{CVPR}
\def\confYear{2024}

\title{Label-free Anomaly Detection in Aerial Agricultural Images with Masked Image Modeling}

\author{
Sambal Shikhar  \hspace{0.3em} Anupam Sobti \\
Plaksha University\\
Mohali, Punjab, India\\
{\tt\small sambal.shikhar@plaksha.edu.in}\\
{\tt\small anupam.sobti@plaksha.edu.in}
}

\begin{document}
\maketitle
\input{sec/0_abstract}
\input{sec/1_intro}
{
    \small
    \bibliographystyle{ieeenat_fullname}
    \bibliography{main}
}


\end{document}

%% file: preamble.tex
\usepackage[dvipsnames]{xcolor}
\usepackage[strict]{changepage}
\usepackage{tabularx} 
\usepackage{booktabs} 


%% file: sec/0_abstract.tex
\begin{abstract}
Detecting various types of stresses (nutritional, water, nitrogen, etc.) in agricultural fields is critical for farmers to ensure maximum productivity. However, stresses show up in different shapes and sizes across different crop types and varieties. Hence, this is posed as an anomaly detection task in agricultural images.
Accurate anomaly detection in agricultural UAV images is vital for early identification of field irregularities. Traditional supervised learning faces challenges in adapting to diverse anomalies, necessitating extensive annotated data. In this work, we overcome this limitation with self-supervised learning using a masked image modeling approach. Masked Autoencoders (MAE) extract meaningful normal features from unlabeled image samples which produces high reconstruction error for the abnormal pixels during reconstruction. 
To remove the need of using only ``normal" data while training, we use an anomaly suppression loss mechanism that effectively minimizes the reconstruction of anomalous pixels and allows the model to learn anomalous areas without explicitly separating ``normal" images for training. 
Evaluation on the Agriculture-Vision data challenge shows a \textbf{6.3\%} mIOU score improvement in comparison to prior state of the art in unsupervised and self-supervised methods. A single model generalizes across all the anomaly categories in the Agri-Vision Challenge Dataset \cite{chiu2020agriculture}.
\end{abstract}

%% file: sec/1_intro.tex
\graphicspath{{./imgs/}}
\section{Introduction}
\label{sec:intro}
\begin{figure}
    \centering
    \includegraphics[width=\linewidth]{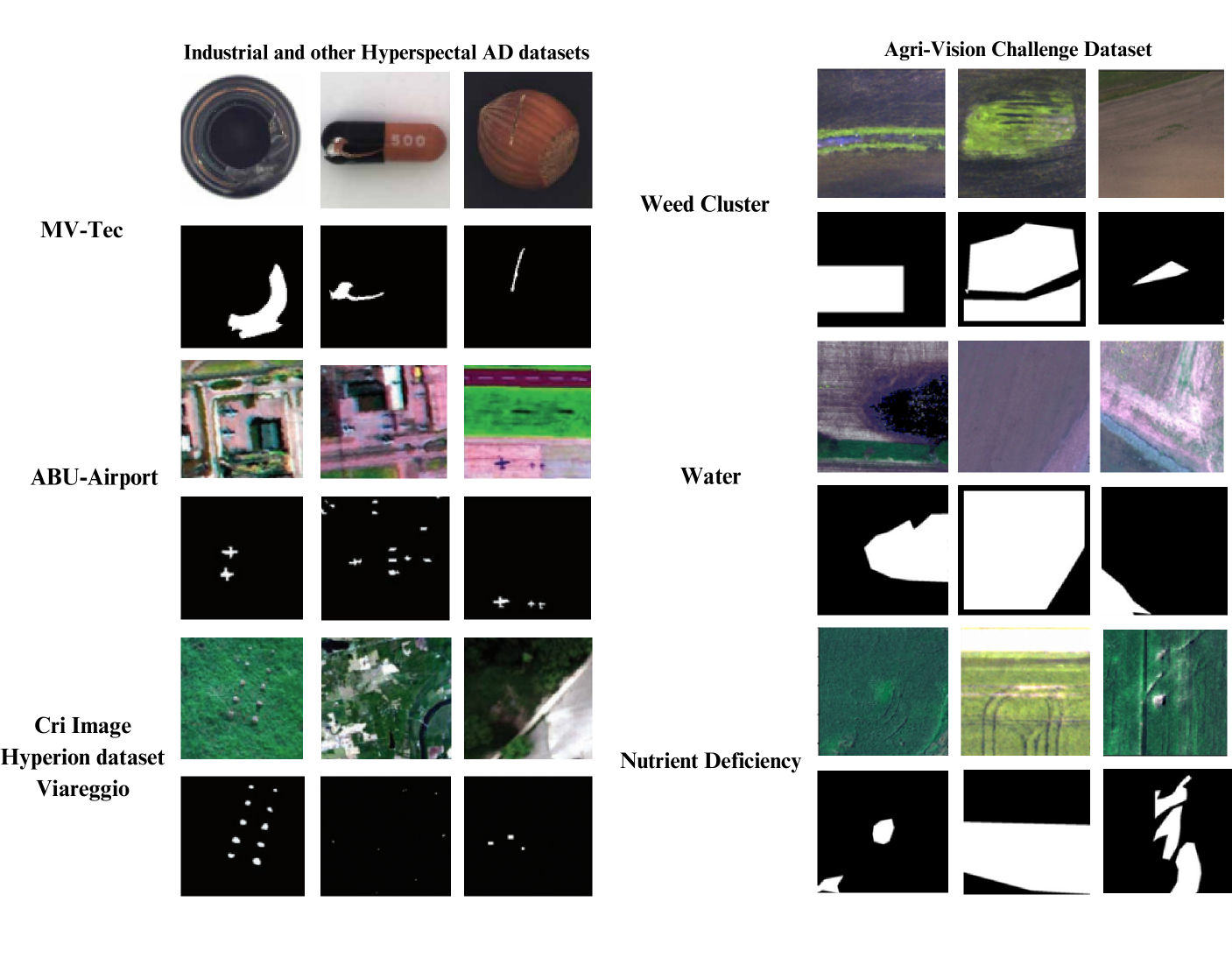}
    \caption{Comparison of anomaly datasets: The left column represents a variety of industrial and other hyperspectral anomaly detection (AD) datasets, including MV-Tec, ABU-Airport, and Cri Image Hyperion dataset of Viareggio. The right column displays the Agri-Vision Challenge Dataset, highlighting agricultural anomalies such as Weed Clusters, Water stress, and Nutrient Deficiency. This illustrates the complexity of agricultural anomalies, showcasing their large inter-class and intra-class variations and their occurrence at multiple scales, as opposed to more uniform and scale-consistent anomalies found in industrial datasets.}
    \label{fig:anomaly_dataset_comparison}
\end{figure}
In precision agriculture, Unmanned Aerial Vehicles (UAVs) have emerged as a pivotal tool for monitoring agricultural landscapes efficiently. UAVs provide much higher resolution images compared to satellite images, thus capturing fine grained details on the agricultural fields.  Accurate anomaly detection in UAV images is crucial for the early identification of potential issues such as pest infestations, diseases, and nutrient deficiencies. 
The dynamic and diverse nature of agricultural fields further compounds the challenge, as anomalies can vary greatly in appearance due to factors such as crop type, growth stage, and environmental conditions as compared to other anomaly detection settings, compared in Figure \ref{fig:anomaly_dataset_comparison}. Thus, there is a need for a completely label free approach to training anomaly detection models so that it can be applied across different crops for different kinds of anomalies. 
Traditionally, \textit{supervised learning} methods have been used for anomaly detection systems \cite{shen2022aaformer,innani2021fuse,chiu20201st,9395478}. These methods are inherently limited by their dependence on large sets of annotated data, which are labor-intensive to create and may not capture the full spectrum of possible anomalies.
Even in case of \textit{unsupervised} and \textit{ self supervised} methods
\cite{chow2020anomaly,zhang2021anomaly,zavrtanik2021reconstruction,liu2021unsupervised,li2021cutpaste} where explicit anomaly labels are not used, there is a dependence on using only ``normal'' data for training thus making it necessary for a user to curate normal data that does not contain any types of anomalies. 

To leverage self-supervised learning through masked image modeling, we utilize Masked Auto-encoders (MAE) \cite{he2022masked} to effectively learn normal features from unlabeled image samples. This ``normality" then facilitates the detection of anomalies through higher reconstruction errors for patches containing anomalies. 
Incorporating a Swin Transformer-based Masked Autoencoder \cite{dai2023swin} enables our model to learn both local and global features, ensuring robust detection across a wide range of anomaly types.
Our work introduces an approach that also learns to detect anomalies with abnormal samples within the training data. This inclusion allows users to simplify their data collection pipeline by removing the need to curate ``normal" data. 
This enables the identification of a wide array of anomalies without training multiple models for detecting different anomalies (by removing anomalies of a particular type from the data).
The effectiveness of our approach is demonstrated through extensive evaluation on the Agriculture-Vision Challenge dataset \cite{chiu2020agriculture}, showcasing significant improvement of 6.3\% in mean Intersection over Union (mIOU) score and generalization across all the given anomaly classes. 
\section{Related Work}

The Agriculture-Vision dataset\cite{chiu2020agriculture} provides multispectral aerial images for fields at 10cm/pix resolution along with annotations for anomalies of 9 types - drydown, planet skip, water, weed cluster, nutrient deficiency, endrow, double plant, waterway, and, storm damage.
Unlike other datasets that may include hyperspectral and multispectral data for general land cover classification \cite{helber2019eurosat} or crop type identification \cite{tseng2021cropharvest}, Agriculture-Vision specifically targets the semantic segmentation of agricultural patterns for recognizing various field anomaly patterns crucial to farmers. 

\textbf{Supervised image segmentation} approaches like FusePN \cite{innani2021fuse} and AAFormer \cite{shen2022aaformer} have demonstrated competitive performance on detecting anomalies in UAV images. AAFormer uses a transformer based architecture. FusePN\cite{innani2021fuse} uses a multimodal fusion approach fusing RGB and NIR bands of the image along in an encoder-decoder style architecture with additional modifications for inference efficiency. The limitations of supervised methods have steered research towards unsupervised and self-supervised learning approaches, where the focus shifts to learning from unlabeled data.

\textbf{One-class classification (OCC)} \cite{tax2004support, zhang2021anomaly} models provide another approach for anomaly detection utilizing high-level semantic information for anomaly identification in the feature space. OCC based anomaly detection uses high-level semantic information and distance metrics for anomaly scoring. However, they encounter challenges such as i) mode collapse and ii) overlook low-level structural features due to their focus on compact feature representation \cite{reiss2021panda}. Anomaly Segmentation based on pixel Descriptors (ASD) \cite{li2023anomaly}
addresses anomaly segmentation in high spatial resolution (HSR) imagery by using deep one-class classification with discriminative pixel descriptors through abnormal sample generation, promoting descriptor compactness for normal data and diversity to prevent model collapse. ASD employs a multi-level, multi-scale feature extraction approach to capture low-level and semantic information.

\textbf{Reconstruction}-based anomaly detection methods like Attribute Restornation Network (ARnet) \cite{ye2020attribute} and Deep Feature Reconstruction (DFR) \cite{yang2020dfr} utilize autoencoders (AE) with an encoder-decoder architecture to capture the manifold of defect-free images to differentiate between normal and anomalous data based on reconstruction fidelity. These models, trained solely on normal imagery, are expected to yield higher reconstruction errors for anomalous inputs, using metrics such as mean square error (MSE) for anomaly quantification. Techniques to enhance anomaly detection capabilities include image degradation and subsequent restoration, notably through inpainting methods like RIAD \cite{zavrtanik2021reconstruction}, which mask parts of the image to challenge the model's reconstruction abilities. Despite their success, these methods struggle, as with progression in training  it inevitably involves the anomalies in the reconstructed image. This is because models favor learning all the information from input, including both background and anomalies simultaneously.
In terms of architectural elements, convolutions are prone to learn identity mapping (from input image to output image) as their receptive fields are biased towards learning local spatial features \cite{liu2023bigset}. To address these limitations, recent approaches propose integrating Transformer based reconstruction method like IntRA (Inpainting Transformer) \cite{pirnay2022inpainting} which pose anomaly detection as a patch-inpainting problem and propose to solve it with a purely self-attention based approach discarding convolutions.Other transformer based approaches like MAE which mask $\sim$75\% of the images patches and use the remaining to reconstruct the complete image. MAEDAY \cite{schwartz2024maeday} leverages MAE for image-reconstruction-based anomaly detection method that utilizes a pre-trained model, enabling its use for Few-Shot Anomaly Detection (FSAD).
We provide a class-wise comparison of these methods in Table \ref{tab:main-benchmark}.

\section{Background}
\textbf{Objective} - The anomaly segmentation of a high spatial resolution (HSR) image \textbf{X} with dimensions \textbf{H×W×B} (height, width, and Number of multi-spectral bands) is defined by a mapping function \textbf{f}, transforming \textbf{X} into an anomaly map\textbf{ A } with dimensions\textbf{ H×W }.

\subsection{Masked Auto-encoder}

A Masked Autoencoder (MAE) proposed by he et al. \cite{he2022masked} learns image representations through self-supervised learning and masked image modelling where a Transformer based model learns to reconstruct an input image by reconstructing an input from a partially masked version of itself.
We use masked image modelling framework and MAE for anomaly detection as given in Figure \ref{fig:Masked auto-encoder for anomaly detection} to learn the background or normal patterns of an input image by reconstructing it from a subset of observed pixels. This process enables the MAE to identify deviations from the learned data distribution, which are indicative of anomalies. A Masked Auto-encoder (MAE) leverages Vision Transformer (ViT) \cite{dosovitskiy2020image} for both its encoder \( E \) and decoder \( D \) components. The ViT segments the input image into patches and applies self-attention mechanisms to capture complex and global features.

Consider an input image \( X \) with dimensions \( H \times W \times B \), where \( H \), \( W \), and \( B \) denote the height, width, and number of bands, respectively. A binary mask \( M \) is applied to generate the masked image \( X_{\text{masked}} \):
\begin{figure*}
    \centering
\includegraphics[width=1\linewidth,height=2\textheight,keepaspectratio]{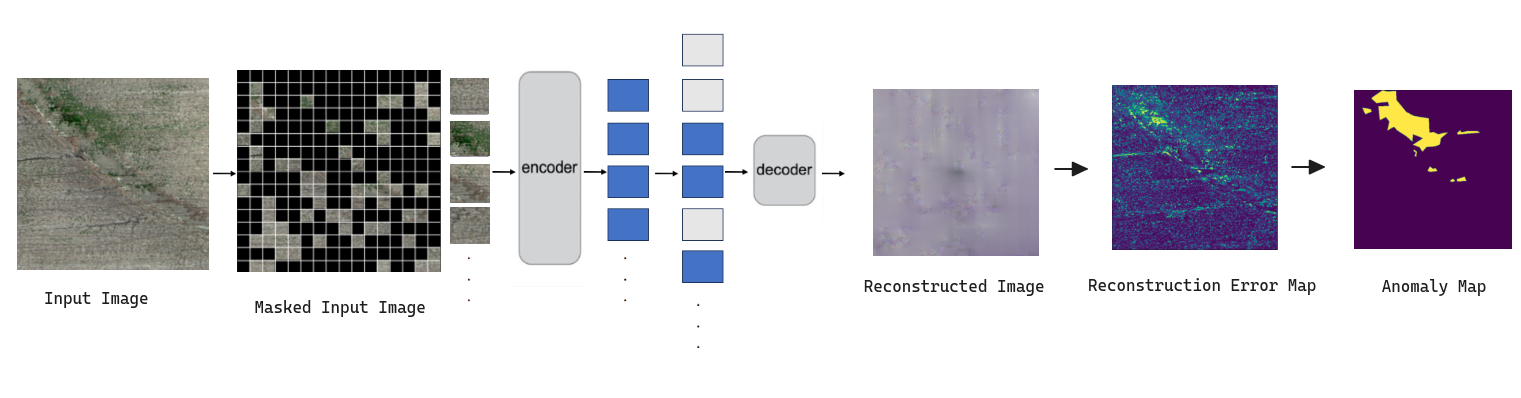}
    \caption{Input image is masked and the unmasked image patches are fed into the encoder which embeds each of those patches ,the decoder takes in embed patches along with masked patches to reconstruct the input image.A reconstruction error map is generated which is then used to generated the final Anomaly map}
    \label{fig:Masked auto-encoder for anomaly detection}
\end{figure*}
\begin{equation}
X_{\text{masked}} = X \odot M
\end{equation}
The encoder \( E \) processes the image by dividing it into \( N \) patches, each with a fixed size \( P \times P \):
\begin{equation}
\text{Patches}_{E} = \text{Patchify}(X_{\text{masked}})
\end{equation}
These patches are then flattened and linearly embedded to a dimension \( D \), followed by adding positional embeddings to retain spatial information:
\begin{equation}
Z_0 = [ \textbf{x}^1_\text{p}E; \textbf{x}^2_\text{p}E; \ldots; \textbf{x}^N_\text{p}E ] + \textbf{E}_\text{pos}
\end{equation}
Here, \( \textbf{x}^i_\text{p}E \) denotes the embedded patches, and \( \textbf{E}_\text{pos} \) is the positional embedding.

The sequence of embeddings \( Z_0 \) is passed through \( L \) Transformer layers to generate the latent representation \( Z_L \):

\begin{equation}
Z_L = \text{Transformer}(Z_{l-1}), \quad \text{for} \ l = 1 \ldots L-1
\end{equation}
Each Transformer layer comprises multi-headed self-attention (MSA) \cite{vaswani2023attention} and multi-layer perceptrons (MLP), with layer normalization (LN) applied before each module and a residual connection after each:
\begin{equation}
Z'_l = \text{MSA}(\text{LN}(Z_{l-1})) + Z_{l-1}
\end{equation}
\begin{equation}
Z_l = \text{MLP}(\text{LN}(Z'_l)) + Z'_l
\end{equation}
The decoder \( D \), structured similarly to \( E \), reconstructs the original image from \( Z_L \):
\begin{equation}
\hat{X} = \text{Patchify}^{-1}(\text{Transformer}_D(Z_L))
\end{equation}
The Transformer layers in \( D \) upsample the latent representations to the original resolution.

The reconstruction error \( E_{\text{recon}} \) between \( X \) and \( \hat{X} \) serves as a measure for anomaly detection:

\begin{equation}
E_{\text{recon}} = ||X - \hat{X}||^2
\end{equation}

This error is evaluated per pixel to generate an anomaly map \( A \) by thresholding the map by $\theta$ :

\begin{equation}
A(i,j) = E_{\text{recon}}(i,j) \geq \theta
\end{equation}

The anomaly detection in MAE is predicated on the assumption that the model, trained predominantly on normal data, will yield higher reconstruction errors for anomalies in \( X \) due to deviations from the learned patterns. The ViT architecture's self-attention mechanism allows the MAE to capture predominantly global features.
\subsection{Swin Transformers}
Swin Transformers \cite{liu2021swin} efficiently handles image representation by partitioning the input image into a grid of patches, which are then processed using self-attention within local windows. The local self-attention mechanism for a patch $P_{i,j}$ is defined as:
\begin{equation}
\text{SA}_{\text{local}}(P_{i,j}) = \text{Softmax}\left(\frac{Q_{i,j} K_{i,j}^T}{\sqrt{d}}\right)V_{i,j}
\end{equation}
where $Q_{i,j}$, $K_{i,j}$, and $V_{i,j}$ are the query, key, and value matrices, respectively, and $d$ is the dimensionality of the query and key.

The Swin Transformer expands the receptive field through a novel shifting mechanism that broadens the scope of self-attention across neighboring patches:
\begin{equation}
\text{Shift}(W_{i,j}) = W_{i+s,j+s}
\end{equation}
where $W_{i,j}$ represents the original window of patches, and $s$ is the shift size.

For multi-scale representation, patches are merged to form larger patches in deeper layers, reducing the resolution while expanding the receptive field:
\begin{equation}
P_{i,j}^{l+1} = \text{Transform}\left([P_{2i,2j}^l, P_{2i+1,2j}^l, P_{2i,2j+1}^l, P_{2i+1,2j+1}^l]\right)
\end{equation}
where $P_{i,j}^l$ denotes a patch at layer $l$, and the Transform function fuses features from four adjacent patches into a new patch at layer $l+1$.

This hierarchical approach enables Swin Transformers to capture global and local patterns and detect anomalies at multiple scales in agricultural fields to enhance their performance when compared to ViT based Masked Autoencoders for anomaly detection tasks.

\begin{figure}
    \centering
\includegraphics[width=0.7\linewidth,height=0.7\textheight,keepaspectratio]{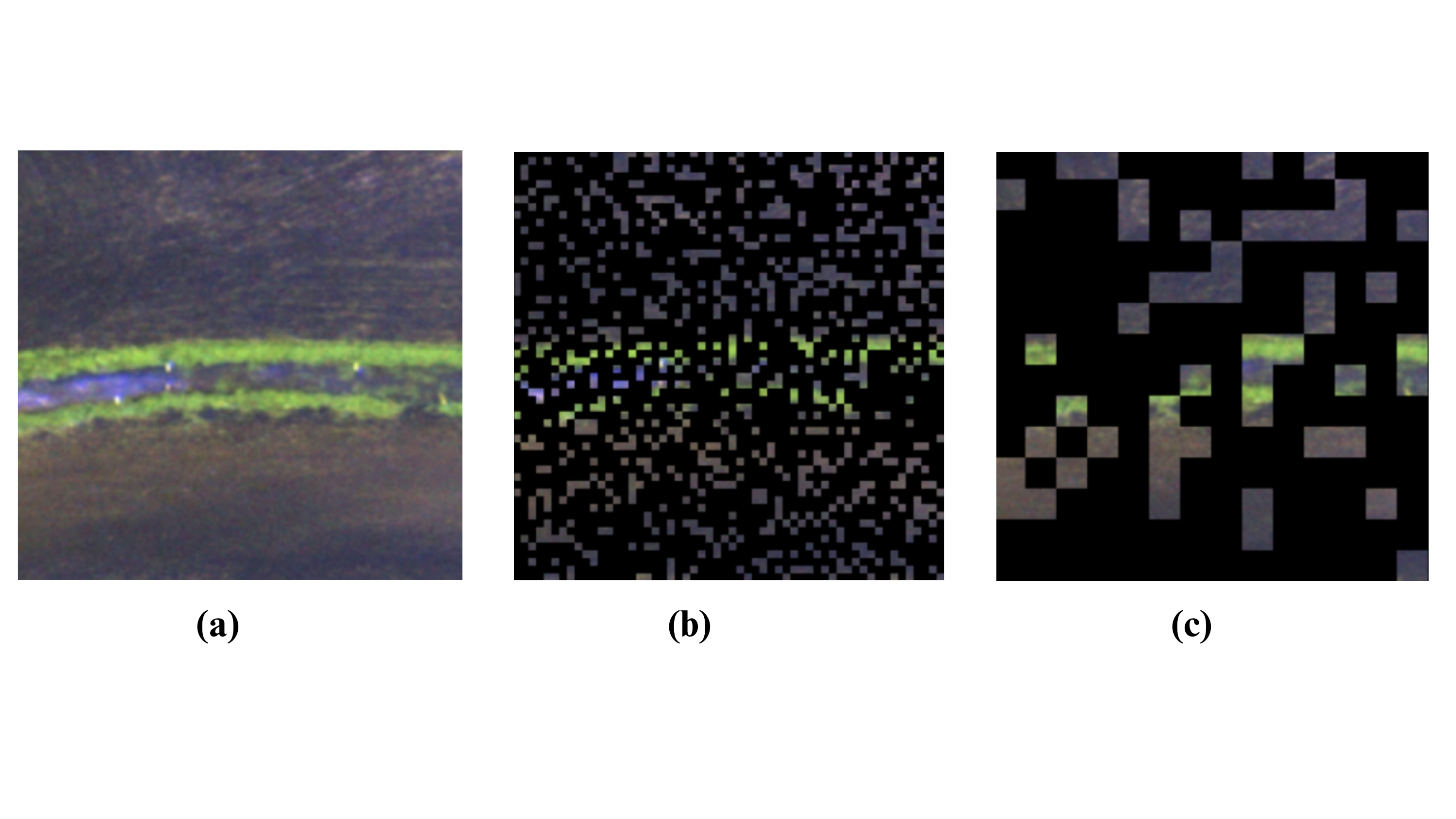}
    \caption{Comparison between masking methods. (a) original
image (b) Normal random masking method (c) Window
masking method.}
    \label{fig:masking}
\end{figure}

\subsection{SwinMAE (Swin Masked Auto-encoder)}
\label{sec:swinmae}
To leverage Swin transformers ability to learn both local and global features and integrate it with masked image modelling framework, Swin Masked Auto-encoder \cite{dai2023swin} architecture replaces the Vision Transformer (ViT) typically used in MAEs with Swin Transformers. The masking strategy in Swin Masked Autoencoder involves a novel approach that maintains the number of patches in the input data during the encoding process whereas MAE only feeds unmasked patches into the encoder. Instead of removing masked patches, which could lead to a shortage of tokens necessary for subsequent processing steps like patch merging, the encoder replaces these masked tokens with a learnable vector. This method ensures a consistent number of tokens throughout the encoding process.

The Swin MAE's window masking strategy addresses the limitations of patch-based masking using MAE, particularly when using smaller patches like 4x4 used in the starting blocks of Swin Transformers as shown in \ref{fig:masking} . This approach divides the image into larger, non-overlapping windows, each containing multiple patches, and masks these windows instead of individual patches. This method aims to prevent models from learning shortcuts, such as reconstructing masked areas through simple interpolation using neighbouring unmasked patches while also maintaining a consistent number of tokens throughout the encoder.

\begin{figure*}
    \centering
\includegraphics[width=0.55\linewidth,height=0.55\textheight,keepaspectratio]{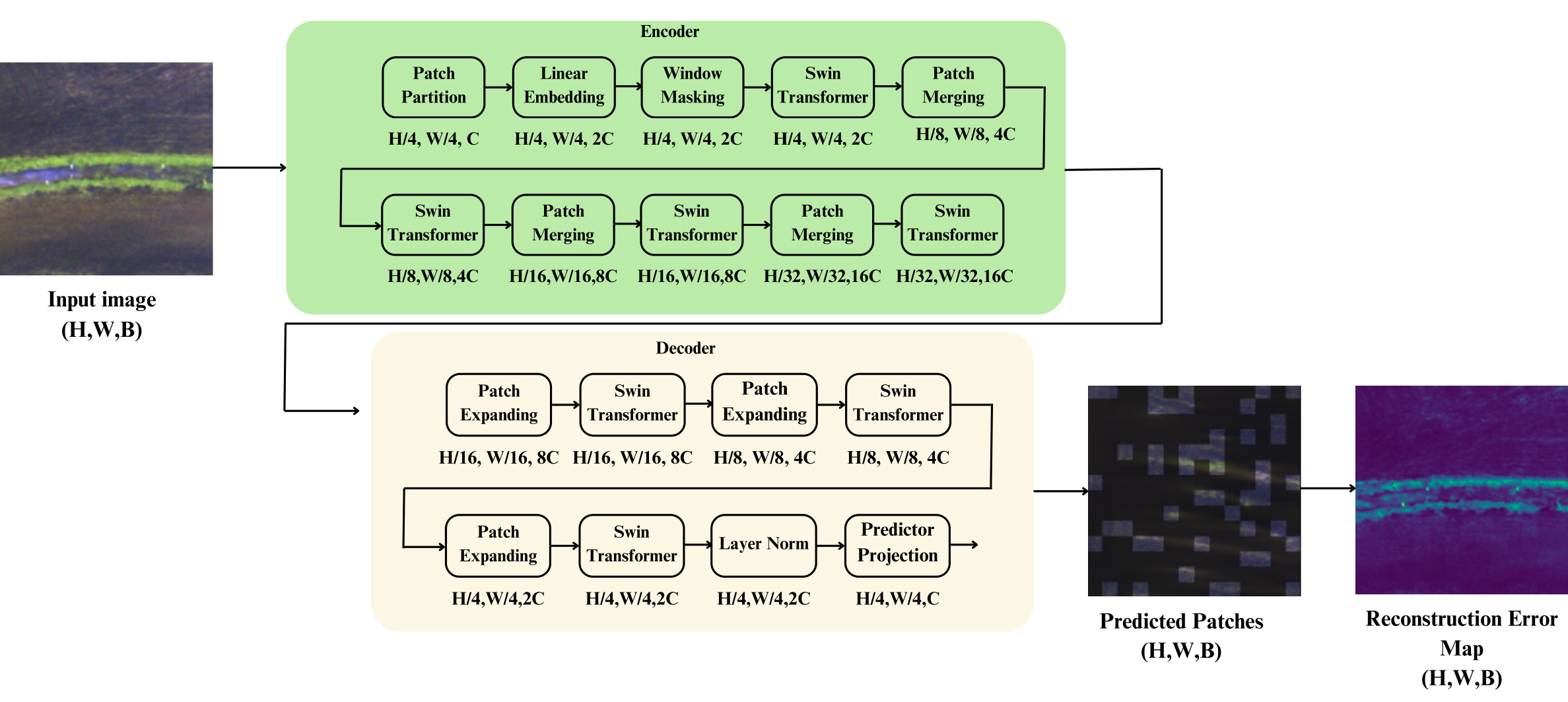}
    \caption{Architecture of the Swin Masked Autoencoder (Swin MAE) for anomaly detection. The encoder, leveraging Swin Transformer blocks, processes the input image through stages of patch partitioning, embedding, and window masking, followed by successive Transformer and merging layers to create high-dimensional token representations. The decoder employs a sequence of expanding, Transformer, and normalization layers before projecting back to the pixel space, resulting in the reconstructed image and its corresponding reconstruction error map.}
    \label{fig:Swin MAE Architecture}
\end{figure*}

Swin MAE uses a light weight decoder with patch expanding layers to restore the image back to its original dimensions which is similar to Swin-Unet \cite{cao2021swinunet} . The decoder consists of Swin transformer blocks. Unlike MAE the masked tokens are not removed through out the encoder so there is no need add these masked tokens in the decoder input. The decoder uses a projection layer to finally restore the image back to its original dimension instead of a patch expanding layer used in Swin-Unet , just like MAE decoder.

\subsection{Anomaly Suppression Loss}
\label{sec:anomaly_supression_loss}
A Mean Square Error (MSE) based loss function allows the model to learn to reconstruct anomaly pixels as the training progresses. During the initial iterations, the reconstruction error for anomalies is higher than for the background pixels, but as more iterations are completed, reconstruction-based models are able to reconstruct the anomaly pixels. Wang et al. \cite{wang2021auto} introduced an adaptive-weighted loss function  which aims to improve anomaly detection by modifying the training focus, emphasizing background pixel reconstruction over anomaly pixels. Note that Wang et.al. demonstrated the method using hyperspectral imagery in a non-agricultural anomaly setup. This method employs a weight map that adjusts the impact of each pixel on the loss based on its reconstruction error, calculated as 
\begin{equation}
e_{i,j} = (x_{i,j} - \tilde{x}_{i,j})^2
\end{equation}
where \( x_{i,j} \) represents the true pixel value, and \( \tilde{x}_{i,j} \) is its reconstruction by the network.
A reconstruction error map \( E \) is constructed from these errors, serving as a basis for anomaly detection:
\begin{equation}
E = \begin{bmatrix}
d_{1,1} & \dots & d_{1,W} \\
\vdots & \ddots & \vdots \\
d_{H,1} & \dots & d_{H,W}
\end{bmatrix}
\end{equation}
To prevent anomalies to get reconstructed, an adaptive-weighted loss function is utilized. The weight map \( W \) is derived by taking residual from the maximum error pixel :
\begin{equation}
w_{i,j} = \max(E) - e_{i,j}
\end{equation}
\begin{equation}
W = \begin{bmatrix}
w_{1,1} & \dots & w_{1,W} \\
\vdots & \ddots & \vdots \\
w_{H,1} & \dots & w_{H,W}
\end{bmatrix}
\end{equation}
\begin{figure*}
    \centering
\includegraphics[width=0.6\linewidth,height=0.6\textheight,keepaspectratio]{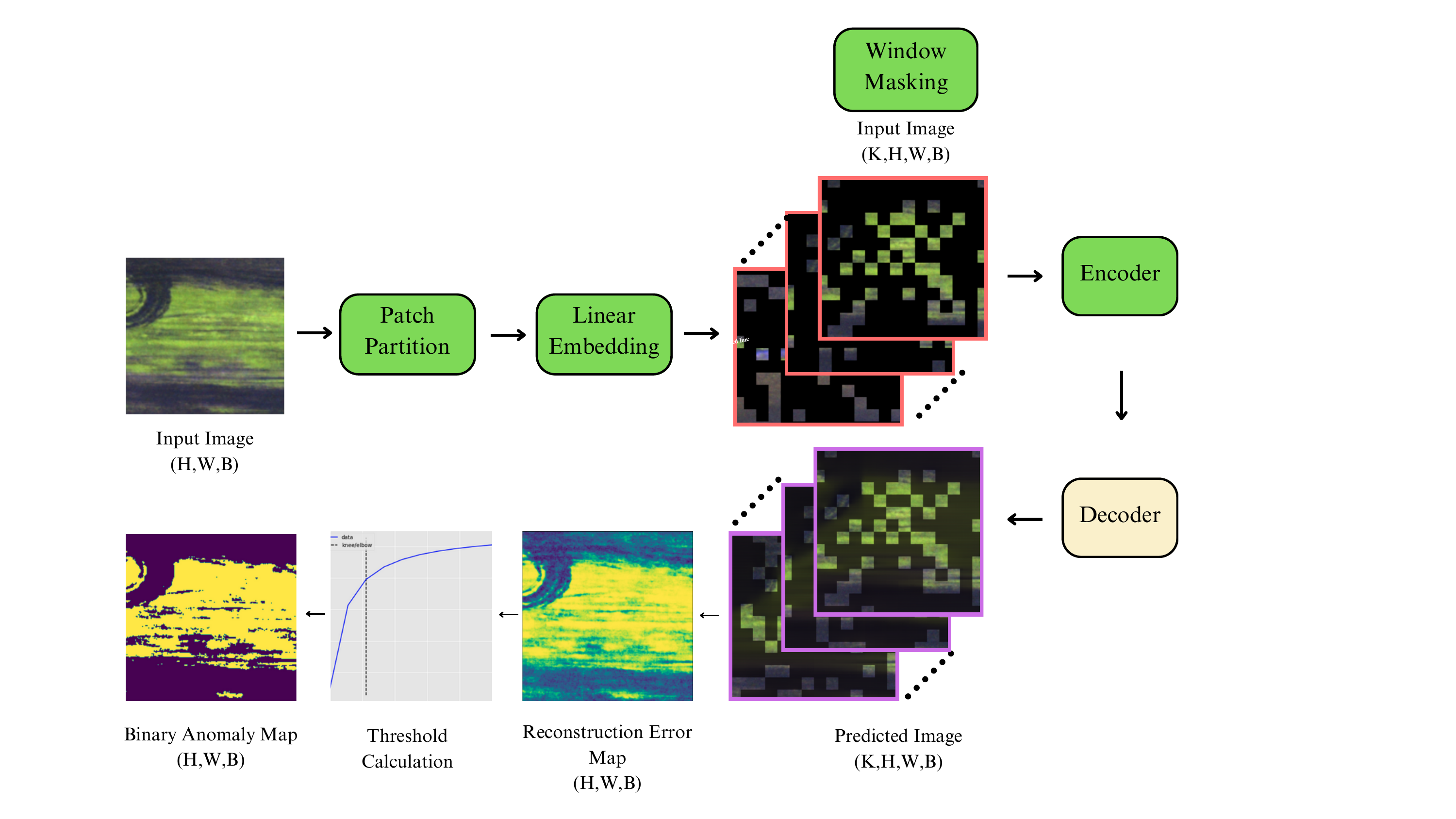}
    \caption{Anomaly Detection using Swin Masked Auto-encoder. An UAV image of shape Height,Width,Number of Bands (H,W,B) is input to the Swin MAE encoder where K window masked images are fed into the rest of the encoder comprised of Swin Tranformer and Patch Merging layers. The Swin MAE decoder then produces resulting K predicted images (K,H,W,B) . The predicted image is compared to the original input image to produce a reconstruction error map, which is thresholded using Knee-point calculation producing the final binary anomaly map delineating the detected anomalies within the image.}
    \label{fig:Swin MAE workflow}
\end{figure*}
After a few initial iterations the reconstruction error for anomalous pixels are very high when compared to background pixels , taking a residual from the maximum error pixel allows anomaly pixels to have less weight compared to the background pixels as the majority of anomaly pixels are closer to the maximum error. This weight map is updated periodically. The adaptive-weighted loss \( L \) is then computed as:

\begin{equation}
L = \sum_{i=1}^{H} \sum_{j=1}^{W} w_{i,j} e_{i,j}
\end{equation}

By reducing the weights of pixels with large reconstruction errors early in training, the network is discouraged from focusing on anomalies, thus prioritizing background reconstruction. This method leads to an anomaly-suppressed model that can more accurately identify anomalies based on the reconstruction error map.

\section{Proposed Method}
We use the SwinMAE architecture (Section \ref{sec:swinmae}) along with the anomaly suppression loss (Section \ref{sec:anomaly_supression_loss}) to learn the ``normal" feature embeddings from the farm images. The architecture of SwinMAE is divided into encoder and decoder as given in Figure \ref{fig:Swin MAE Architecture}.
The encoder of the Swin MAE network begins by partitioning the input image into non-overlapping patches and mapping these patches into a high-dimensional embedding space through a linear transformation, which allows for more complex feature extraction.The patches then go through widow masking strategy. The Swin Transformer blocks, which form the core of the encoder effectively captures hierarchical features by using window based multi-head self-attention (W-MSA) followed by shifted (SW-MSA) window mutli-head attention. After each Swin Transformer block the embeddings are merged to reduce the number of patches by half while doubling the dimensionality of the embeddings. This operation aggregates information from adjacent patches and reduces the spatial resolution, while increasing the feature dimension. The encoder consists of 4 Swin transformer blocks as given in the original work \cite{dai2023swin}.
The decoder in the Masked Autoencoder (MAE) network undertakes the task of reconstructing the input image from its condensed and partially masked representation produced by the encoder. It begins by processing the mixed embeddings through Swin Transformer blocks. Following this, the decoder employs a series of expanding operations that gradually restore the spatial resolution of the image. This is essentially the reverse of the encoder's patch merging process. As the resolution is increased, the complexity of the feature representation is reduced, aligning it closer to the original input space. Layer normalization steps interspersed within these operations ensure stable learning by maintaining a consistent scale of the features. The final stage involves a projection of the embeddings back to a space that mirrors the original image's patches, which are then reassembled to predict the full image, effectively filling in the masked regions with the learned information. This reconstructed output aims to be as close as possible to the original unmasked image. 


Given an input image \( X \), a subset of patches \( P \) representing 25\% of \( X \) are fed into the SwinMAE model as given in Figure \ref{fig:Swin MAE workflow}. This process is repeated \( K \) times with different random subsets, where \( K = 32 \) so that each patch is likely to be masked once, allowing for accurate reconstruction assessment for each pixel.For each iteration \( i \), a reconstruction \( R_i \) is obtained and compared with \( X \) to compute a reconstruction error map \( E_i \). The error for each pixel \( j \) in the error maps is averaged across all \( N \) reconstructions to obtain an averaged error map \( \bar{E} \):
\begin{equation}
\bar{E}_j = \frac{1}{K} \sum_{i=1}^{K} E_{ij}
\end{equation}
The final anomaly map \( A \) is produced by applying a threshold \( \theta \), determined by identifying the knee point \cite{satopaa2011finding} in the distribution of \( \bar{E} \), to binarize \( \bar{E} \) into anomalous (\( A_j = 1 \)) and non-anomalous (\( A_j = 0 \)) pixels:
\begin{equation}
A_j = 
\begin{cases} 
1 & \text{if } \bar{E}_j \geq \theta \\
0 & \text{otherwise}
\end{cases}
\end{equation}
\begin{table*}
\centering
\setlength{\tabcolsep}{2pt} 
\renewcommand{\arraystretch}{1.2} 
\begin{tabularx}{\textwidth}{@{}l *{10}{X}@{}} 
\toprule[1.5pt]
Method & Drydown & Double plant & Endrow & Weed cluster & ND & Water & Planter Skip & Waterway & Storm Damage & mIOU* \\ 
\midrule[0.75pt]
DSVDD\cite{zhang2021anomaly} & 30.8 & 14.7 & 10.1 & 3.0 & 24.7 & 26.0 & -  & -  & - & 18.2  \\
RIAD\cite{zavrtanik2021reconstruction} & \textbf{31.0} & 25.6 & 27.3 & 38.2 & 25.7 & \textbf{42.7}  & -  & -  & - &  31.7 \\
ARNet\cite{ye2020attribute} & 30.6 & 15.3 & 25.5 & 15.9 & 26.4 & 9.6 & - & -  & - & 20.5 \\
GANomaly\cite{akcay2019ganomaly} & 26.3 & 4.2 & 26.4 & \textbf{41.9} & 33.6 & 20.7& -  & -  & - &  25.3 \\
ASD\cite{li2023anomaly} & 34.6 & 24.8 & 25.7 & 19.7 & 31.7 & 40.4 & -  & -  & - &  29.4 \\
InTra\cite{pirnay2022inpainting} & 26.1 & 44.2 & 41/0 & 36.8 & 33.4 & 35.2 & 51.0  & \textbf{47.1}  & \textbf{42.5} &  29.3 \\
MAE & 27.6 & 44.0 & 43.1  & 34.6 & 33.2 & 35.2 & 50.2  & 45.7  & 41.7 & 36.2 \\
SwinMAE & 27.9 & 46.3 & 43.3 & 36.7 & 33.8 & 37.4 & 51.5  & 45.3  & 41.9  & 37.5 \\ 
SwinMAE + ASL (Ours) & 28.1 &\textbf{ 46.6} & \textbf{43.8} & 37.8 & \textbf{34.1} & 37.8 & \textbf{52.7}  & 46.9  & 42.3  & \textbf{38.0}\\ 
\bottomrule[1.5pt]
\end{tabularx}
\caption{The comparative performance of anomaly segmentation methods on the Agriculture-Vision dataset. ASL = Anomaly Suppression Loss. SwinMAE based anomaly detection is able to beat existing benchmarks. Anomaly Suppression Loss (ASL) further improves accuracy through selective suppression of anomalies. *Since previous methods reported mean across 6 classes, we also reported mean IoU over 6 classes for a fair comparison. These numbers are reported by training in a leave-one-out fashion.}
\label{tab:main-benchmark}
\end{table*}
\section{Experimental Setup}
\subsection{Dataset}

We evaluate anomaly detection on Agriculture Vision challenge dataset.The Agriculture-Vision dataset \cite{chiu2020agriculture} is a large aerial image database aimed at agricultural pattern analysis, designed to boost research in computer vision for agriculture. It contains 94,986 high-quality aerial images from 3,432 farmlands across the U.S., with each image including RGB and Near-infrared (NIR) channels with resolutions up to 10 cm per pixel.The dataset employs a cropping technique on large farm images with a window size of 512 × 512 pixels for annotations.The images are annotated with 9 types of field anomaly patterns such as such as double plant, drydown, endrow, nutrient deficiency, water,weed cluster,planter skip , storm damage and waterway which are crucial to farmers and serves as a benchmark for agricultural semantic segmentation, posing unique challenges due to the large inter-class and intra-class variations.
A total of 56,944 images for training, 18,334 for validation and 19,708 for testing is created. We benchmark our methods on all 9 classes of anomalies, while previous results were only reported on 6 classes - Double Plant, Drydown, Endrow, Nutrient Deficiency, Water and Weed Cluster.

\subsection{Evaluation Metrics}

We evaulate the anomaly detection task a semantic segmentation task and use Intersection over Union (IoU), as it quantitatively assess how closely the predicted anomaly map aligns with the ground truth annotation.
\\
\begin{equation}
\text{IoU} = \frac{\text{TP}}{\text{TP} + \text{FP} + \text{FN}}
\end{equation}

\begin{equation}
\text{mIoU} = \frac{1}{N} \sum_{i=1}^{N} \frac{TP_i}{TP_i + FP_i + FN_i}
\end{equation}

\begin{figure*}[ht]
    \centering
    \begin{subfigure}[b]{0.48\linewidth}
        \centering
        \includegraphics[width=\linewidth,keepaspectratio]{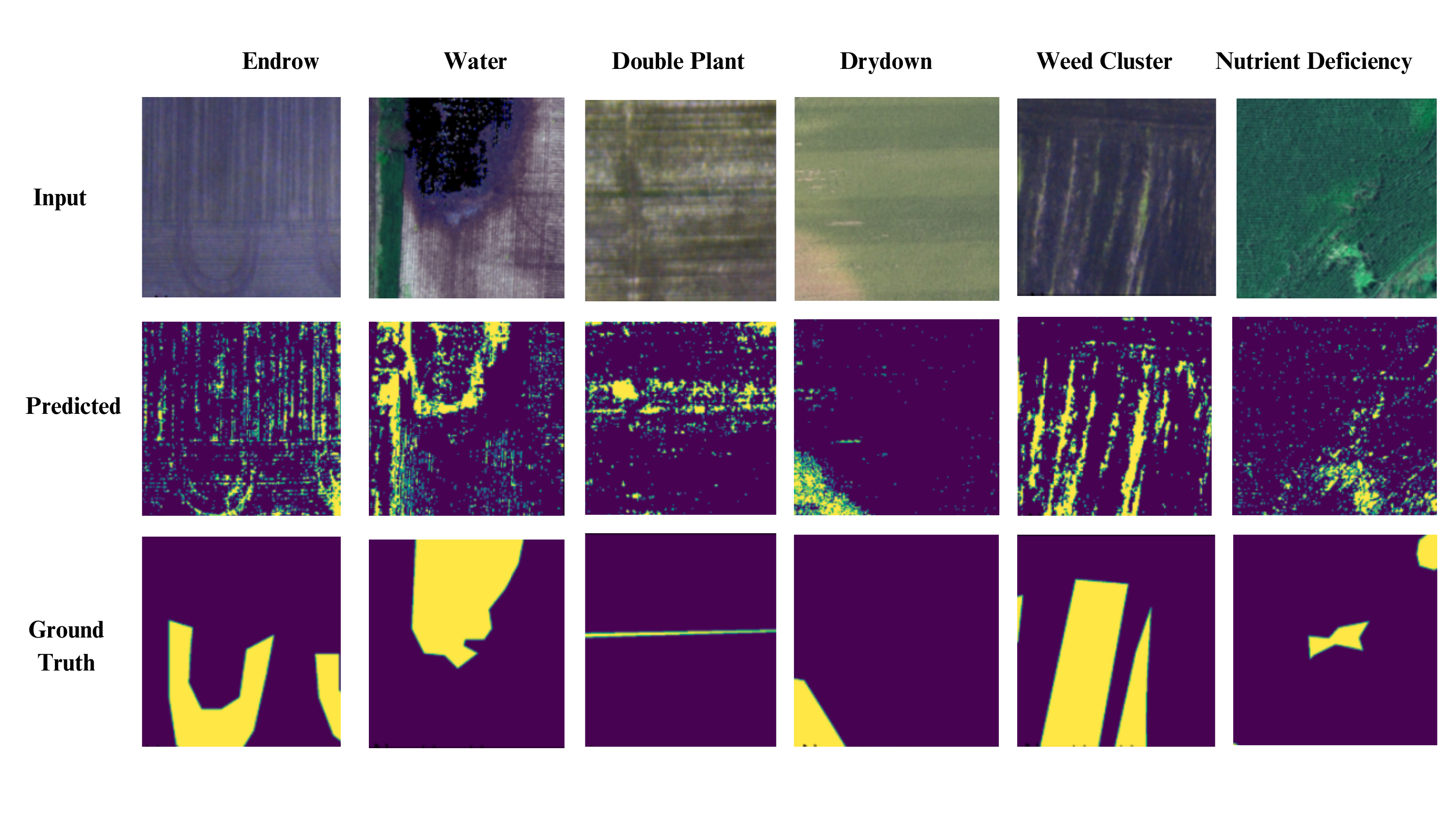}
        \caption{Good examples}
        \label{fig:correct_qualitative_results}
    \end{subfigure}%
    \hfill
    \begin{subfigure}[b]{0.48\linewidth}
        \centering
        \includegraphics[width=\linewidth,keepaspectratio]{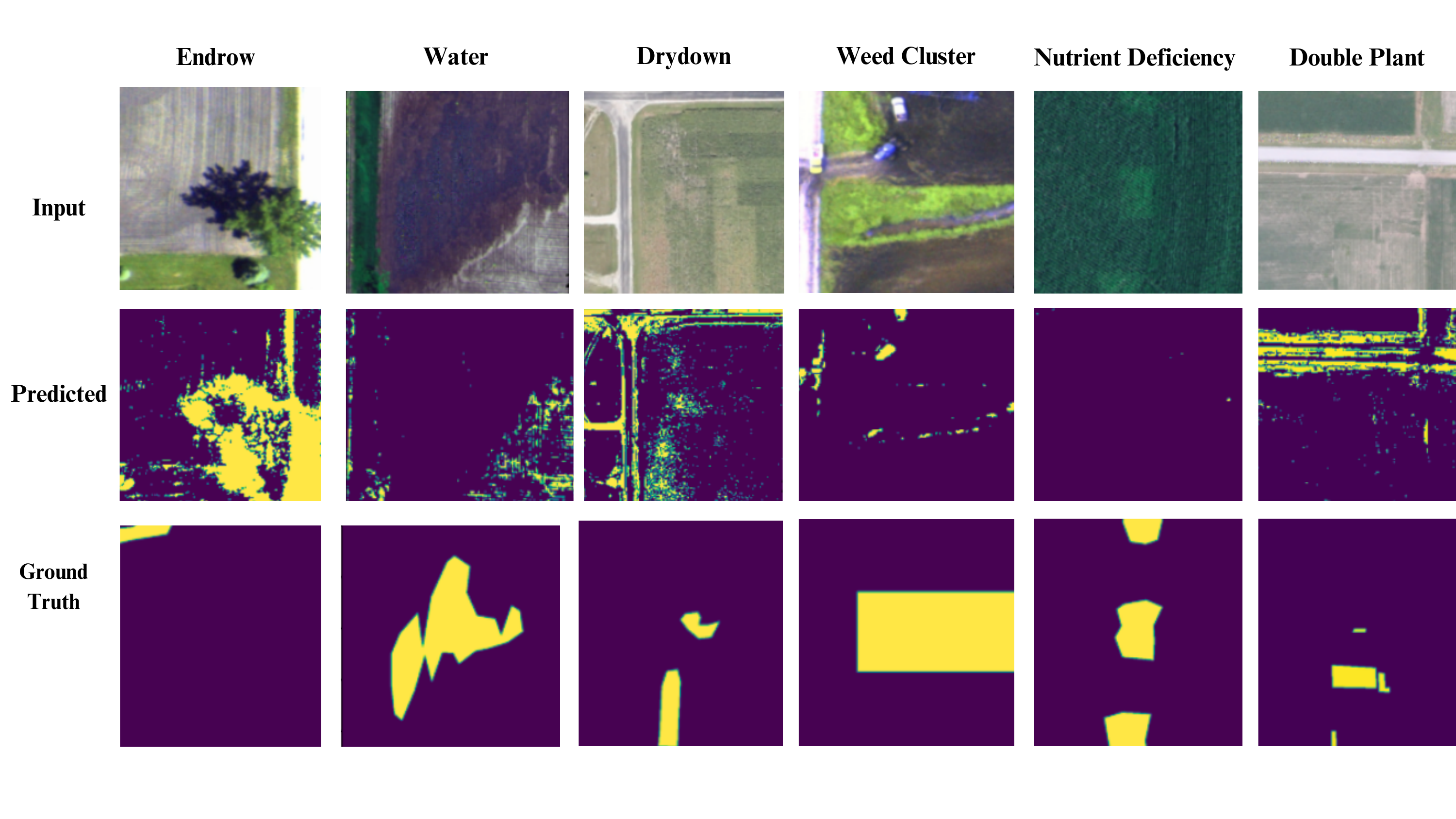}
        \caption{Poor Examples}
        \label{fig:incorrect_qualitative_results}
    \end{subfigure}
    \caption{Qualitative results demonstrating the model's performance across six anomaly classes.}
    \label{fig:qualitative_results}
\end{figure*}

\subsection{Implementation Details}

The input images to SwinMAE are resized from 512x512 to 224x224 for computational efficiency and 4 bands were used, RGB and Near-infrared (NIR). The SwinMAE takes in the entire training data as input across all the classes and provides a single model for anomaly detection across the given classes.The encoder and decoder comprises of 4 Swin Transformer blocks.The initial input image patch size is 4x4.
AdamW \cite{yao2021adahessian} optimizer with initial learning rate of 1e-3 and weight decay was used to train the model. Experiments were conducted on a single 24G A5000 GPU and on a machine with 256GB RAM.The model was trained on a total of 200 epochs and with a batch size of 64.
We train SwinMAE for 20 epochs initially and then the weight maps are updated for anomaly loss compression. Also, to benchmark MAE, the training setup remains the same as SwinMAE .

\begin{table*}[ht]
\centering
\begin{tabularx}{\textwidth}{@{}l *{11}{X}@{}} 
\toprule[1.5pt]
Training Data & DryD & DP & Endrow & WC & ND & Water & PSkip & WW & SD & mIOU & medIOU\\ 
\midrule[0.75pt]
Excluding anomalous samples & \textbf{45.83} & \textbf{ 46.9} & 43.5 & \textbf{37.9} & 33.7 & \textbf{45.97}  &\textbf{ 53.3 }&\textbf{ 47.5} & 42.0 & \textbf{44.0} & \textbf{45.8} \\
Including anomalous samples & 28.1 & 46.6 & \textbf{43.8} & 37.8 &\textbf{ 34.1} & 37.8 & 52.7 & 46.9 &\textbf{ 42.3} & 36.44 & 42.3 \\
\bottomrule[1.5pt]
\end{tabularx}
\caption{When anomalous samples are included in the dataset, the anomaly suppression loss helps maintain the IoU over most classes. DryDown and Water show losses since these are semantically very different from the input data distribution and spread widely in the image. The median is very close to the original paradigm while the mean is lower due to the two classes - Drydown and Water.}
\label{tab:secondary-benchmark}
\end{table*}

\section{Results}

%

We compare SwinMAE with several state of the art anomaly detection algorithms ranging from convolutional, GAN-based, One-class classification (OCC), Transformers and MAE based models. To be specific quantitative analysis of Swin MAE in comparison to several models such as DSVDD \cite{zhang2021anomaly}, RIAD \cite{zavrtanik2021reconstruction}, ARNet \cite{ye2020attribute}, GANomaly \cite{akcay2019ganomaly}, and the Anomaly Segmentation model based on Pixel Descriptors (ASD) \cite{li2023anomaly}, Inpainting Transformer \cite{pirnay2022inpainting} and MAE.
Our quantitative experiments in Table \ref{tab:main-benchmark} reveals that Swin MAE outperforms previous unsupervised and self supervised approaches across several anomaly categories. Supervised models like AAFormer \cite{shen2022aaformer} achieves an mIOU of 41.2 and Fuse-PN \cite{innani2021fuse} achieves a dice score of 82.71 acting as strong baselines. For instance, in the  Double plant and Endrow category, Swin MAE achieved an mIOU of 46.6 and 43.8 which is significantly better than other models. Swin MAE performs on par if not better across all the categories of anomaly. Moreover, when augmented with an anomaly suppression loss, the model's proficiency further increases, where it recorded a jump in all categories showing the efficacy of the anomaly suppression mechanism during training. A Masked autoencoder also performs well compared to the other models , which enforces the fact that masked image modelling is very effective in detecting anomalies across multiple categories. However as Swin MAE learns both local and global features it outperforms MAE in all the categories. It is interesting to note that anomalies which spread out in the images, e.g., weeds, water, etc. are not caught well with reconstruction based methods, perhaps, owing to the fact that despite 75\% masking, the pattern is caught by the encoder and successfully reproduced, thus acting as the normal distribution.

\textbf{Label-free training} We also analyze training SwinMAE with and without anomaly samples in the training set for a given class, as detailed in Table \ref{tab:secondary-benchmark}. Note that all methods in Table \ref{tab:main-benchmark} are run without anomalies in the training data, i.e., in a Leave-one-out fashion. In general, it is observed that with our method, accuracies remain almost the same (or increase slightly by a few points in some cases) for all classes of anomalies. The median IOU only reduces by 3.5 points. For two classes - drydown and water, the IoU increases significantly when these classes are omitted during training. This effect can be attributed to the distinctive nature of these two classes, i.e., water and drydown not being present at all in the underlying data, which are less likely to be reconstructed if they are absent from the training set due to their minimal distribution overlap with other types of anomaly classes which are present in the training set. 

These results indicate that Swin MAE is robust across most of the anomaly categories while it is trained with anomalous samples included in the training set and significantly outperforms other models which are trained on anomaly free samples which enables us to have to a single model better generalizing to all categories. Swin MAE ability to model both local and global dependencies across patches enables it to detect anomalies better than other transformer based techniques like a simple MAE and InTra.

Qualitatively we can observe in Figure \ref{fig:correct_qualitative_results}, that SwinMAE is able to segment irregular anomaly patterns across multiple classes. Due to large intra-class and inter-class variance within every class in the Agriculture Vision Dataset \cite{chiu2020agriculture}, we observe that in a few classes like Nutrient Deficiency as shown in Figure \ref{fig:correct_qualitative_results} there are other miscellaneous anomaly patterns apart from nutrient deficient areas that get segmented as anomalies, similarly for classes like water and double plant. In Figure \ref{fig:incorrect_qualitative_results} there are objects such as Trees in (Endrow), Roads in (Drydown, Double Plant) and cars in (Weed) that are detected as anomalies because they deviate from normal agricultural patterns and global pattern of the image, however these are not agricultural anomalies. As our model is trained in a self-supervised setting and without agricultural anomaly specific supervision, any pattern that deviates from the global setting of the query image might be labeled as anomaly. This also solidifies the fact that Masked image modelling based transformer models are efficient global feature learners.

\section{Conclusion and Future Work}
We propose a masked image modelling based self supervision methodology to detect anomalies in agricultural fields using UAV images. We demonstrate the effectiveness of masked image modelling through SwinMAE\cite{dai2023swin} and MAE\cite{he2022masked} methods by benchmarking it on the Agriculture Vision dataset and improving mIOU by a margin of 6.3\% compared to other unsupervised and self supervised methods. We also show that using an anomaly suppression loss\cite{wang2021auto} adds robustness even when trained with training data containing anomalous samples. Typically, the anomaly detection methods train only on normal samples (excluding the anomalous classes' patches from the input). This improvement allows a simplification of the pipeline so that anomaly detection pipelines can be trained without filtering out ``normal" samples. With this improved methodology, it is also possible to have a single model that can generalize across all the anomaly classes. Our work should provide a definitive direction in creating an anomaly detection system that generalizes to various anomalies and relies less on human interventions.
As future work, we will investigate if the single model is able to find ``new" anomalies beyond the one described in the dataset and if the method is useful in domains other than UAV images as well.  We will also investigate if a generic model will be able to work across different types of crops.